\RequirePackage{fix-cm}

\documentclass{svjour3}                     

\smartqed  

\usepackage{graphicx}

\usepackage{amsmath}
\usepackage{url}
\usepackage{multirow}

\usepackage{subcaption}
\usepackage{enumerate}
\usepackage[inline, shortlabels]{enumitem}

\begin{document}

\title{Video-based surgical skill assessment using 3D~convolutional neural networks
}


\author{Isabel Funke \and S{\"o}ren Torge Mees \and J{\"u}rgen Weitz \and Stefanie Speidel}


\institute{I. Funke \and S. Speidel \at
Division of Translational Surgical Oncology, National Center for Tumor Diseases (NCT), \mbox{Partner} Site Dresden, Dresden, Germany \\
\email{Firstname.Lastname@nct-dresden.de}
           \and
S. T. Mees \and J. Weitz \at
Department of Visceral, Thoracic and Vascular Surgery, Faculty of Medicine and University Hospital Carl Gustav Carus, TU Dresden, Dresden, Germany
}

\date{Received: date / Accepted: date}

\maketitle

\begin{abstract}
~ \smallskip \\ 
\noindent
\emph{Purpose}: A profound education of novice surgeons is crucial to ensure that surgical interventions are effective and safe. One important aspect is the teaching of technical skills for minimally invasive or robot-assisted procedures. This includes the objective and preferably automatic assessment of surgical skill. 

Recent studies presented good results for automatic, objective skill evaluation by collecting and analyzing motion data such as trajectories of surgical instruments. However, obtaining the motion data generally requires additional equipment for instrument tracking or the availability of a robotic surgery system to capture kinematic data. In contrast, we investigate a method for automatic, objective skill assessment that requires video data only. This has the advantage that video can be collected effortlessly during minimally invasive and robot-assisted training scenarios.

\noindent
\emph{Methods}: Our method builds on recent advances in deep learning-based video classification. Specifically, we propose to use an inflated 3D~ConvNet to classify 
snippets, i.e., stacks of a few consecutive frames, extracted from surgical video.
The network is extended into a Temporal Segment Network during training. 

\noindent
\emph{Results}: We evaluate the method on the publicly available JIGSAWS dataset, which consists of recordings of basic robot-assisted surgery tasks performed on a dry lab bench-top model.
Our approach achieves high skill classification accuracies ranging from 95.1\% to 100.0\%.

\noindent
\emph{Conclusions}: Our results demonstrate the feasibility of deep learning-based assessment of technical skill from surgical video.
Notably, the 3D~ConvNet is able to learn meaningful patterns directly from the data, alleviating the need for manual feature engineering.  
Further evaluation will require more annotated data for training and testing. 

\end{abstract}

\section{Introduction}
\label{sec:intro}

A profound education of novice surgeons is crucial to increase patient safety and to avoid adverse operation outcomes. This holds especially for technically challenging surgical techniques, such as traditional and robot-assisted minimally invasive surgery.
One important aspect of teaching surgery is \emph{surgical skill assessment}. It is a prerequisite for targeted feedback, which facilitates efficient skill acquisition by telling the novice how to improve. Additionally, skill assessment is important for certifying prospective surgeons.

Traditionally, novice surgeons are taught and evaluated by experts in the field. To ensure objective and standardized assessments, procedure-specific checklists and global rating scales for surgical skill evaluation were introduced~\cite{Ahmed2011}. However, these evaluation protocols make skill assessment very time-consuming and thus expensive.
For this reason, automating the process of surgical skill evaluation is a promising idea. Besides saving time and money, this would enable novice surgeons to train effectively in absence of a human supervisor using a surgical simulator that is equipped with automatic assessment and feedback functionalities.
Consequently, \emph{objective computer-aided technical skill evaluation (OCASE-T)}~\cite{Vedula2017} has received increasing interest in the research community over the last decade.    

For computer-aided technical skill evaluation, sensor data is being collected during surgical training, for example video data, robot kinematics, tool motion data, or data acquired by force and torque sensors.
This data is then analyzed in order to estimate the surgical skill of the trainee. Here, surgical skill can either be described numerical, for example as average OSATS~\cite{Martin1997} score, or categorical, for example using the classes novice, intermediate, and expert.

One approach for OCASE-T is to calculate descriptive metrics from the observed data and to use these metrics to determine surgical skill. For example, Fard et al.~\cite{Fard2018} record instrument trajectories and extract several features from the trajectories, such as path length, motion smoothness, curvature, and task completion time. Using these features, they train various machine learning classifiers
 to distinguish between expert and novice.
Zia and Essa~\cite{Zia2018} evaluate texture features, frequency-based features, and entropy-based features extracted from robot kinematic data. For classifying surgeons into expert, intermediate, or novice, they employ a nearest neighbor classifier after dimensionality reduction using principal component analysis. They achieve very high classification accuracies using a feature based on \emph{approximate entropy (ApEn)}.

Another approach is to model the collected data as time series using Markov Models, Hidden Markov Models, or variations thereof. For example, Tao et al.~\cite{Tao2012} propose a \emph{Sparse Hidden Markov Model (S-HMM)}.
For surgical skill classification, one S-HMM per class is learned from performances of the respective skill level. For classifying a new performance, the model that generates this perfomance with the highest likelihood is determined and the corresponding skill class is returned.

Recently, also methods based on Convolutional Neural Networks were proposed.
\emph{Convolutional Neural Networks (ConvNets)} are able to learn hierarchical representations from data. They can serve as feature extractors that build high-level features from low-level ones. 
As implied by recent studies~\cite{Fawaz2018,Wang2018}, they can also learn to detect discriminative motion patterns in robot kinematic data in order to distinguish accurately between expert, intermediate, and novice surgeons.

A brief introduction to ConvNets and deep learning is given in~\cite{Lecun2015}.
Notably, ConvNets can process data in their raw form. In contrast, conventional machine learning classifiers require data to be transformed into an adequate representation, such as the motion features described above. Such features usually need to be designed manually, which can be a cumbersome process that depends on experience in feature engineering and requires domain knowledge. 

Most methods for OCASE-T focus on the analysis of robot kinematic data or tool motion data. However, obtaining this data generally requires the availability of a robotic surgical system or of specialized tracking systems~\cite{Chmarra2007}. 
On the other hand, video data of the surgical field can be easily acquired in traditional and robot-assisted minimally invasive surgery using the laparoscopic camera. However, video data is high-dimensional and considerably more complex than sequences of a few motion variables. To deal with these challenges, approaches working on video data usually apply one of two strategies. 

One strategy is to track the surgical tools in the video and then analyze the obtained tool motion data. There exist a number of algorithms to perform vision-based surgical tool tracking~\cite{Bouget2017}. Also ConvNets have been shown to work well for this task~\cite{Du2018,Laina2017}. In one study, Jin et al.~\cite{Jin2018} use region-based ConvNets for instrument detection and localization. Then, they assess surgical skill by analyzing tool usage patterns, movement range, and economy of motion.

The other strategy is to transform the video data into a $D$-dimensional time series using a bag of words (BoW) approach. For example, Zia et al.~\cite{Zia2018Video} extract spatio-temporal interest points (STIPs) from each video frame. Each STIP is represented by a descriptor composed of a histogram of optical flow (HOF) and a histogram of oriented gradients (HOG). A dictionary of visual words is learned from two distinct expert videos by clustering the STIP descriptors of their video frames using k-means clustering with k $= D$. 
For the remaining videos, each frame is represented by a $D$-dimensional histogram, where the $\text{i}^{\text{th}}$ histogram bin counts the number of STIP descriptors extracted from the frame that belong to the $\text{i}^{\text{th}}$~cluster represented by the $\text{i}^{\text{th}}$~visual word. The $D$-dimensional time series corresponding to one video can then be analyzed using descriptive metrics or time series analysis as described before.

In contrast to both strategies, we propose to process video data without any of the intermediate steps discussed above. Using a 3D~ConvNet, we directly  learn feature representations from video data. This way, we avoid the process of manually engineering suitable intermediate video representations for video-based surgical skill assessment.
To the best of our knowledge, we are the first to present a deep learning-based approach for automatic, objective surgical skill classification using video data only.

To achieve this, we bring together three advances from video-based human action recognition: 
\begin{enumerate*}[label=(\roman*)]
\item a modern 3D~ConvNet architecture~\cite{Carreira2017} to extract spatiotemporal features from video snippets with a typical length of 16~--~64 frames,
\item \mbox{Kinetics}~\cite{Kay2017}, a very large, publicly available human action dataset, which we exploit for cross-domain pretraining, and
\item the Temporal Segment Network (TSN) framework~\cite{Wang2018TSN} to learn to resolve ambiguities in single video snippets by considering several snippets at a time, which are selected using a segment-based sampling scheme.
\end{enumerate*}  
The source code will be made available at \url{https://gitlab.com/nct_tso_public/surgical_skill_classification}.

We evaluate the method on the JIGSAWS dataset~\cite{Ahmidi2017,Gao2014}, which consists of recordings of three elementary tasks of robot-assisted surgery. The tasks are performed in a simulated environment using dry lab bench-top models.
While the JIGSAWS dataset does not contain intraoperative recordings and hence does not represent real surgical scenes, it representes surgical training scenarios very well.
It is common that surgical novices practice on dry lab bench-top trainers,
where they perform tasks of curricula such as the Fundamentals of Laparoscopic Surgery (FLS) \cite{peters2004} or the Structured Fundamental Inanimate Robotic Skills Tasks (FIRST) \cite{goh2015}.
These training scenarios are the ones where we expect automatic skill assessment to have an impact first,
by providing automatic feedback to the novice and thus enabling him or her to practice without human expert supervision.
In general, however, our method can be applied to video recordings of any kind of surgical task as long as a sufficiently large database of labeled videos is available.

Most similar to our approach is the work by Doughty et al.~\cite{Doughty2017} on skill-based ranking of instructional videos. They train a Temporal Segment Network with a modified loss function to learn which video of a pair of videos exhibits a higher level of skill.
However, they do not investigate the use of 3D~ConvNets and therefore rely on spatial image features only.

\section{Methods}
\label{sec:methods}

\begin{figure*}
  \includegraphics[width=0.8\textwidth]{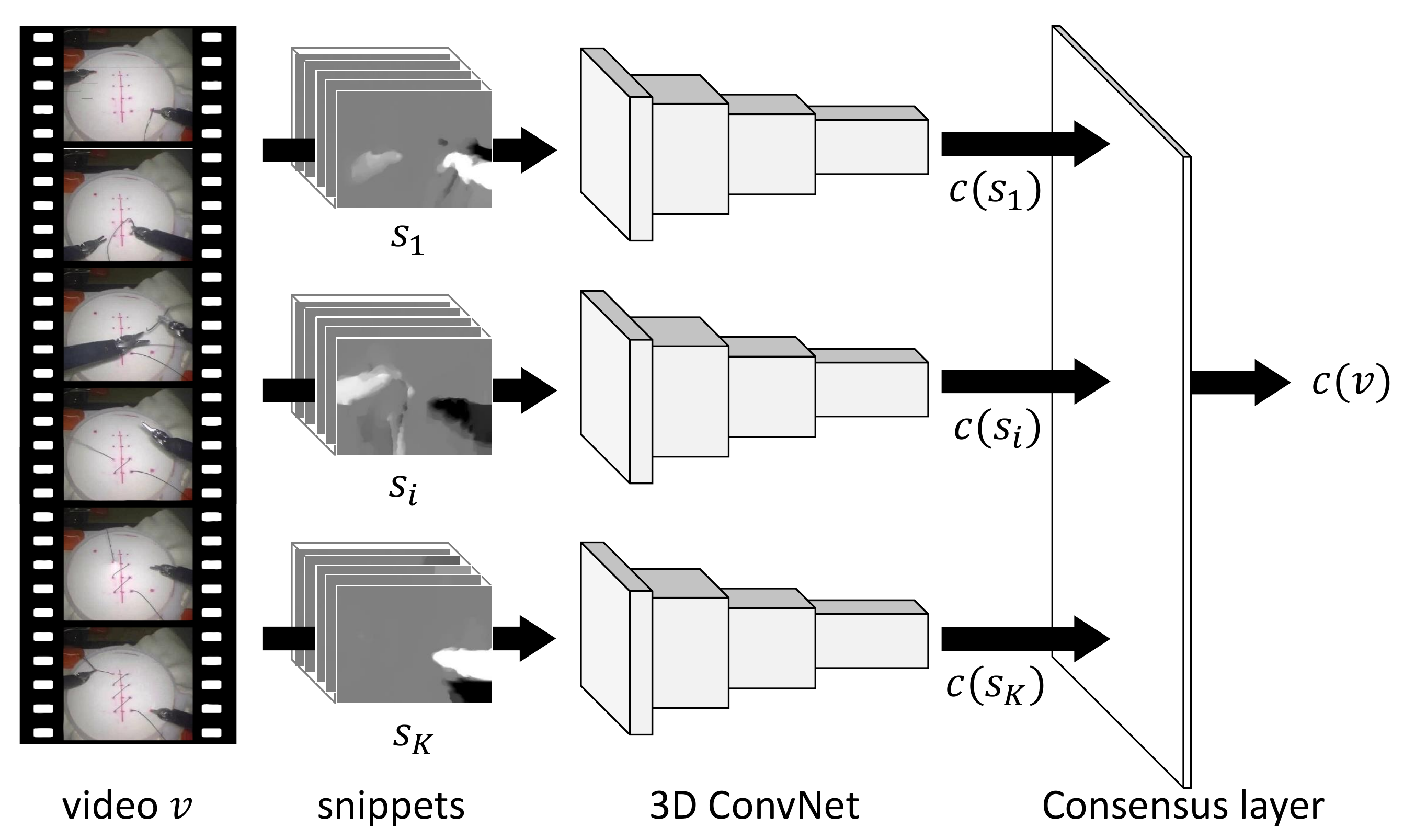}
  \centering
 \caption{The Temporal Segment Network employed for video classification. Each of the $K$ video snippets is classified using one ConvNet instance. The snippet-level classifications $c(s_i)$ are aggregated in the consensus layer, yielding the overall video classification~$c(v)$. Snippets are stacks of several consecutive video frames or, as depicted, stacks of several optical flow fields calculated between consecutive video frames.}
 \label{fig:tsn}       
\end{figure*}

Formally, we solve a video classification problem. Given are a number of videos recorded during surgical training. Each video displays the performance of one surgical task, such as suturing or knot tying. Such a video has a typical length of one to five minutes. The problem is to classify each video as expert, intermediate, or novice depending on the level of surgical skill exhibited in the video. 
 
\emph{Temporal Segment Networks~(TSNs)} were introduced by Wang et al.~\cite{Wang2018TSN,Wang2016} as a deep learning-framework for video classification in the context of human action recognition. The main idea is to use a ConvNet to classify several short snippets taken from a video. Then, the snippet-level predictions are aggregated using a consensus function to obtain the overall video classification result (see figure~\ref{fig:tsn}).
Several differentiable aggregation functions can be used to form consensus. One of the most straightforward choices is average pooling, which has been shown to work well in practice~\cite{Wang2018TSN}. 

The sampling strategy for obtaining the video snippets differs slightly between the training and the testing stage of the TSN. During training, snippets are sampled \emph{segment-based}. This means that each video is partitioned into $K$ non-overlapping segments along the temporal axis. Then, from each segment a snippet consisting of one or more consecutive frames is sampled at a random temporal position within the segment.
This yields $K$ snippets in total.
Usually, $K$ is in the range of 3~to~9 for classifying video clips with a typical length of around 10~seconds~\cite{Wang2018TSN}.
During testing, $\kappa$ snippets are sampled equidistantly from the test video. Usually, the number $\kappa$ is larger than $K$, for example $\kappa = 25$.

During training, a TSN can be seen as a Siamese~\cite{Bromley1994} architecture: several instances of the same ConvNet, which share all of their weights, are used to classify one video snippet each. After forming consensus, the video classification error is backpropagated to update the network parameters. 
This way, the ConvNet learns to classify video snippets in a way that, if being applied to several snippets of a video, the consensus over all predictions yields the correct video class. Notably, forming consensus allows for compensation of contradictions and ambiguities in single video snippets.

A TSN is a promising choice for surgical skill classification from video. Looking at short snippets instead of complete videos is an effective strategy to reduce the size and complexity of the problem at hand. Additionally, the segment-based snippet sampling scheme at training time already realizes some kind of data augmentation. This is appealing because the available training data for the problem of surgical skill assessment is still very limited. 
Moreover, with a TSN it is possible to classify single snippets within the same video different from each other as long as the snippet-level predictions add up to the correct overall result. This is important for video-based skill classification, where usually only an overall assessment per video is given, while individual parts of the video could exhibit different levels of surgical skill. 

In order to implement a TSN for surgical skill classification, two design decisions must be made. First, we need to decide how a single snippet is constructed from a video. Secondly, we need to decide about the architecture of the ConvNet used for snippet classification.

\subsection{Snippet structure}

In the original paper by Wang et al.~\cite{Wang2016}, a snippet is a single RGB video frame or a stack of 5 dense \emph{optical flow~(OF)} fields calculated between consecutive video frames. 
Here, each optical flow frame has two channels: one containing the horizontal component, the other containing the vertical component of optical flow.

Because we want to provide as much information as possible to the network, we decided to use larger snippets consisting of 64~consecutive frames. 
We extract frames from the video at a frequency of 10~Hz,
which means that a snippet of 64~frames represents more than six seconds of surgical video.

While prior work~\cite{Simonyan2014} indicates that the optical flow modality yields better results than the RGB modality for deep learning-based human action recognition, optical flow is also expensive to calculate. Also, optical flow explicitly captures motion between video frames but does not represent the appearance of static objects in the scene.
In this study, we evaluate both modalities. Therefore, a video snippet consists of either 64~consecutive RGB frames or of 64~frames of optical flow calculated between consecutive video frames.

\subsection{ConvNet architecture}
\label{ss:ConvNet}

Originally, Wang et al.~\cite{Wang2018TSN} proposed to use deep ConvNet architectures that are well-established for image classification tasks. Examples are Inception~\cite{Szegedy2016} and ResNet~\cite{He2016}.
However, image classification networks only extract features along the spatial dimensions. They do not model the temporal information encoded in stacks of consecutive video or optical flow frames.

In order to facilitate spatiotemporal feature learning for action recognition, 3D~ConvNets were proposed~\cite{Ji2013,Tran2015}.
In a nutshell, 
3D~ConvNets are a variant of Convolutional Neural Networks with the property that convolutional filters, pooling operations, and feature maps have a third -- the temporal -- dimension.

Carreira and Zisserman~\cite{Carreira2017} describe how to canonically inflate deep 2D~ConvNet architectures along the temporal dimension to obtain deep 3D~ConvNets.
They propose \emph{Inception-v1~I3D}, the 3D~version of Inception.
Because this architecture achieves state-of-the-art results on many human action datasets, we decided to use this network for surgical skill classification.

In order to adapt the network to our problem, we set the number of output neurons to three. 
The output neurons are meant to indicate the surgical skill class, namely expert, intermediate, or novice, using a one-hot encoding.

\subsection{ConvNet pretraining}

To successfully train a 3D~ConvNet from scratch, a lot of training data and a lot of computational resources are required, both of which we are lacking.
For image classification problems, a popular strategy for coping with limited training data is to pretrain the ConvNet on a large labeled dataset coming from a related domain and to finetune it afterwards. 
Fortunately, it has been shown that this also works for 3D~ConvNets and video classification problems~\cite{Carreira2017}. Moreover, pretrained 3D~ConvNet models have become publicly available and are ready to be finetuned on custom datasets.
We chose to use an Inception-v1~I3D model pretrained on the Kinetics dataset, which is publicly available\footnote{\url{https://github.com/deepmind/kinetics-i3d/tree/master/data/checkpoints}} for both input modalities, RGB and optical flow. 

\emph{Kinetics}~\cite{Kay2017} is one of the largest human action datasets available so far. It consists of 400~action classes and contains at least 400~video clips per class.
We assume that the spatiotemporal features learned from the Kinetics dataset are a helpful initialization for subsequently finetuning the 3D~ConvNet for surgical skill classification.

Before finetuning the pretrained model, we freeze the weights of all layers up to, but not including, the second to last Inception block (\texttt{Mixed\_5b}). This way, the number of trainable parameters is greatly reduced, which enables us to finetune the model using one GPU with only 8~GB of video RAM.

\subsection{Implementation and training details}
\label{ss:Details}

We implement the TSN for surgical skill classification using the PyTorch package~\cite{Paszke2017}. Our implementation is based on the source code published with the TSN paper\footnote{\url{https://github.com/yjxiong/tsn-pytorch}} and the PyTorch implementation of Inception-v1~I3D provided by Piergiovanni\footnote{\url{https://github.com/piergiaj/pytorch-i3d}}.
The optical flow fields are calculated using the TV-L1~algorithm~\cite{Zach2007} implemented in OpenCV~\cite{opencv_library}. 
For data augmentation, we use corner cropping, scale jittering, and horizontal flipping as described in~\cite{Wang2016}.
The final width and height of each frame is $224$~pixels, which is the expected input size of \mbox{Inception-v1}~I3D. 

When finetuning the pretrained 3D~ConvNet for surgical skill classification, the parameters of the 
modified output layer
are initialized with random values sampled uniformly from the range $(-\sqrt{\frac{1}{1024}}, \sqrt{\frac{1}{1024}})$.
The dropout probability of the dropout layer, which is located just before the output layer in Inception-v1~I3D, is set to~$0.7$. We use standard cross entropy loss and the Adam optimizer~\cite{Kingma2015} with a learning rate of~$10^{-5}$. We train the TSN for 1200~epochs. During each epoch of training, we sample $K = 10$ snippets from each video in the training set. The batch size is set to~2. For testing, we use $\kappa = 25$ snippets per video.
All hyperparameters were chosen heuristically.

\section{Evaluation}
\label{sec:evaluation}

\begin{figure*}
    \centering
    \begin{subfigure}[t]{0.32\linewidth}
        \centering
        \includegraphics[height=1.2in]{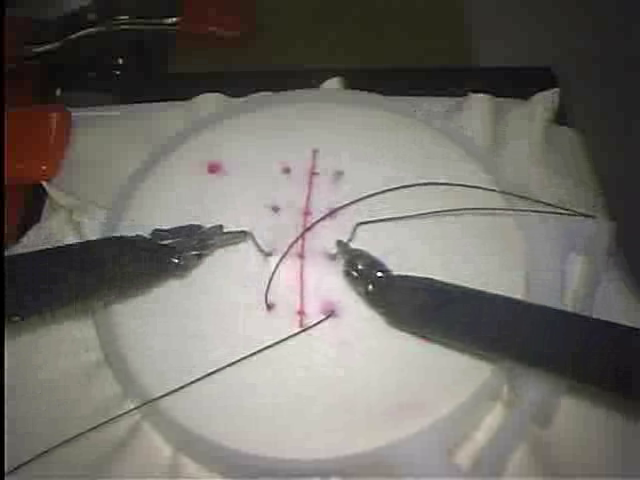}
        \caption{Suturing}
    \end{subfigure}
    \begin{subfigure}[t]{0.32\linewidth}
        \centering
        \includegraphics[height=1.2in]{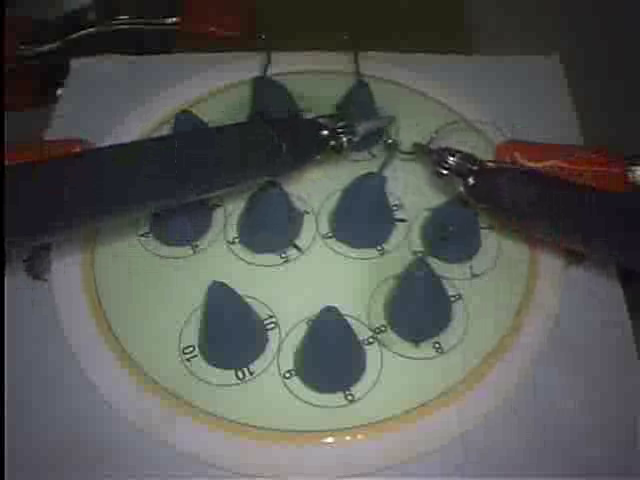}
        \caption{Needle passing}
    \end{subfigure}
    \begin{subfigure}[t]{0.32\linewidth}
        \centering
        \includegraphics[height=1.2in]{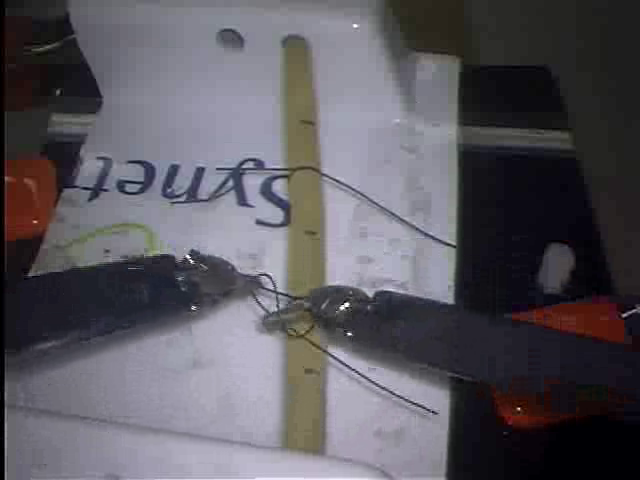}
        \caption{Knot tying}
    \end{subfigure}
    \caption{Surgical tasks recorded in JIGSAWS~\cite{Ahmidi2017}}
    \label{fig:tasks}
\end{figure*}

We evaluate the TSN for surgical skill classification on the \emph{JHU-ISI Gesture and Skill Assessment Working Set (JIGSAWS)}~\cite{Ahmidi2017,Gao2014}, one of the largest public datasets for surgical skill evaluation. It contains recordings of three basic surgical tasks performed on a bench-top model using the da Vinci Surgical System: suturing, needle passing, and knot tying (see figure~\ref{fig:tasks}). 
Each recording contains the synchronized robot kinematics in addition to the laparoscopic stereo video captured during the performance of one task.
Eight participants with varying robotic surgical experience performed each of the tasks up to five times. The participants are classified into the categories expert, intermediate, and novice based on their hours of robotic surgical experience. 

In this work, our goal is to predict the level of surgical experience (expert, intermediate, or novice) on the JIGSAWS dataset using the video data only. The video data in JIGSAWS was recorded at 30~Hz with a resolution of $640 \times 480$ pixels. We only use the right video from each stereo recording. 

The authors define two cross-validation schemes for JIGSAWS: \emph{Leave-one-supertrial-out (LOSO)} and \emph{leave-one-user-out (LOUO)}. For LOSO, the data of one surgical task is split into five folds, where the $\text{i}^{\text{th}}$ fold contains the $\text{i}^{\text{th}}$ performance of each participant. 
For LOUO, the data is split into eight folds, where each fold contains the performances of one specific participant.

\subsection{LOSO evaluation}
\label{ss:loso}

\begin{table}
\caption{Surgical skill classification results on the suturing and needle passing tasks, averaged over four evaluation runs. All measures are given in~\%.}
\label{tab:1}   
\centering   
\begin{tabular}{lllllll}
\hline\noalign{\medskip}
\multirow{2}{*}{Method}  	& \multicolumn{3}{c}{Suturing} & \multicolumn{3}{c}{Needle passing} \\
	 						& accuracy & avg. recall & avg. $F_1$ & accuracy & avg. recall & avg. $F_1$ \\
\noalign{\smallskip}\hline\noalign{\smallskip}
3D ConvNet (RGB)				& $100 \pm 0$ & $100 \pm 0$ & $100 \pm 0$ & $96.4 \pm 0$ & $96.3 \pm 0$ & $96.6 \pm 0$ \\
3D ConvNet (OF)				& $100 \pm 0$ & $100 \pm 0$ & $100 \pm 0$ & $100 \pm 0$ & $100 \pm 0$ & $100 \pm 0$ \\
ConvNet~\cite{Wang2018}		& 94.1 & --- & 92.33 & 90.3 & --- & 87.0\\
ConvNet~\cite{Fawaz2018} 	& 100 & 100 & --- & 100 & 100 & --- \\
ApEn~\cite{Zia2018}  		& 100 & --- & --- & 100 & --- & --- \\
S-HMM~\cite{Tao2012} 		& 97.4 & --- & --- & 96.2 & --- & ---\\
\noalign{\smallskip}\hline
\end{tabular}
\end{table}

\begin{table}
\caption{Surgical skill classification results on the knot tying task, averaged over four evaluation runs. All measures are given in~\%.}
\label{tab:2}   
\centering   
\begin{tabular}{llll}
\hline\noalign{\medskip}
\multirow{2}{*}{Method}  	& \multicolumn{3}{c}{Knot tying} \\
							& accuracy & avg. recall & avg. $F_1$ \\
\noalign{\smallskip}\hline\noalign{\smallskip}
3D ConvNet (RGB)				& $95.8 \pm 1.6$ & $95.6 \pm 1.2$ & $95.9 \pm 1.5$ \\
3D ConvNet (OF)				& $95.1 \pm 2.7$ & $94.2 \pm 3.2$ & $95.0 \pm 2.9$ \\
ConvNet~\cite{Wang2018}		& 86.8 & --- & 82.33\\
ConvNet~\cite{Fawaz2018} 	& 92.1 & 93.2 & ---\\
ApEn~\cite{Zia2018}  		& 99.9 & --- & ---\\
S-HMM~\cite{Tao2012} 		& 94.4 & --- & ---\\
\noalign{\smallskip}\hline
\bigskip
\end{tabular}
\end{table}
 
\begin{figure*}
  \centering
    \begin{subfigure}{0.45\textwidth}
    \includegraphics[width=0.8\textwidth]{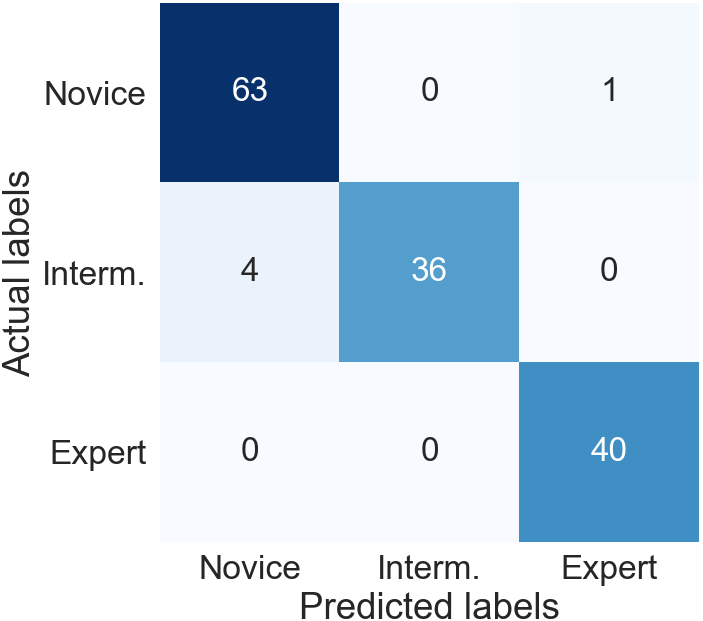}
    \caption{3D ConvNet (RGB)}
    \label{fig:CM1_RGB}
  \end{subfigure}
  \begin{subfigure}{0.45\textwidth}
    \includegraphics[width=0.8\textwidth]{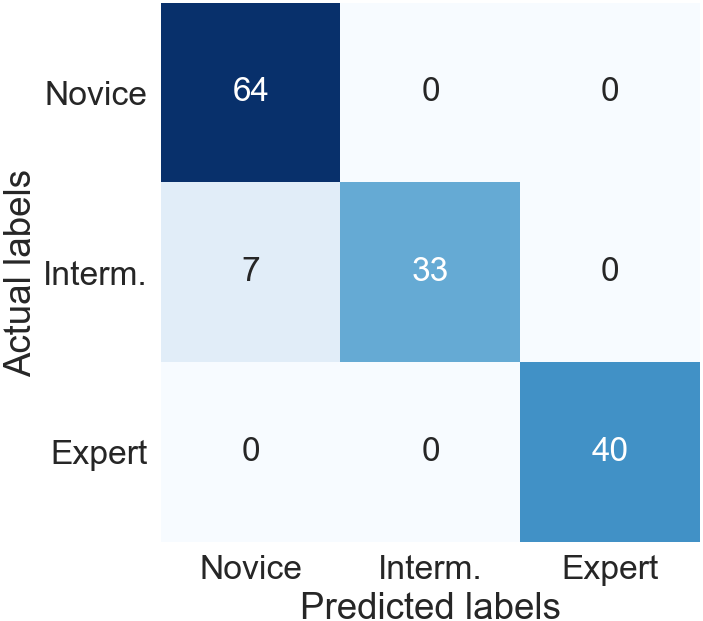}
    \caption{3D ConvNet (OF)}
    \label{fig:CM1_OF}
  \end{subfigure}
   \caption{Predictions on the knot tying task, accumulated over four evaluation runs}
   \label{fig:CM1}
\end{figure*}

We evaluate our approach separately for each of the three surgical tasks, following the LOSO cross-validation scheme.
During one evaluation run for one task, we train a model for each of the five LOSO folds using data from the other folds. Then, we predict the surgical skill class for each video using the model that has not seen this video at training time.

For the obtained predictions, we calculate accuracy, average recall, and average $F_1$~score. \emph{Accuracy} is the proportion of correct predictions among all predictions. Regarding a single class, \emph{recall} describes how many of the instances of this class are recognized as instance of this class. On the other hand, \emph{precision} describes how many of the examples that are predicted to be instance of this class actually are instances of this class. $F_1$~is the harmonic mean of precision and recall. Recall and $F_1$ score are averaged over the three classes expert, intermediate, and novice.
Notably, accuracy and average recall correspond to the \emph{Micro} and \emph{Macro} measures defined in~\cite{Ahmidi2017}. 

We perform our experiments using either RGB or optical flow as input modality.
To account for the stochastic nature of ConvNet initialization and training, we perform four evaluation runs per task and input modality and average the results.
We report mean and standard deviation.
The results can be found in tables~\ref{tab:1} and~\ref{tab:2}, where our method is referred to as \emph{3D~ConvNet~($\ast$)}. 
The input modality employed, namely plain video~(RGB) or optical flow~(OF), is denoted in brackets.

For comparison, we state the results of some state-of-the-art methods described in section~\ref{sec:intro}. Notably, all of these methods utilize the robot kinematics provided in JIGSAWS. Unfortunately, we could not find another video-based method for surgical skill classification that has been evaluated on JIGSAWS.

The confusion matrices in figure~\ref{fig:CM1} visualize the predictions on the knot tying task, using either RGB~(\ref{fig:CM1_RGB}) or optical flow~(\ref{fig:CM1_OF}) as input modality. Here, for matrix~$C$, the number in matrix cell $C(i, j)$, $i, j \in \lbrace 0, 1, 2\rbrace$, denotes how many times an instance of class $i$ is classified as instance of class $j$. Class~$0$ corresponds to skill level \emph{novice}, class~$1$ to \emph{intermediate}, and class~$2$ to \emph{expert}.

\subsection{Ablation studies}

\begin{table}
\caption{Surgical skill classification results of the ablation experiments \ref{abl:3D} -- \ref{abl:Kin} on the knot tying task using the optical flow modality, averaged over four evaluation runs. All measures are given in~\%.}
\label{tab:abl}      
\centering
\begin{tabular}{llll}
\hline\noalign{\medskip}
\multirow{2}{*}{Method}  	& \multicolumn{3}{c}{Knot tying} \\
							& accuracy & avg. recall & avg. $F_1$ \\
\noalign{\smallskip}\hline\noalign{\smallskip}
3D ConvNet (as proposed)	& $95.1 \pm 2.7$ & $94.2 \pm 3.2$ & $95.0 \pm 2.9$ \\
Using 2D ConvNet~\ref{abl:3D}		& $87.5 \pm 1.6$ & $85.6 \pm 2.1$ & $85.5 \pm 2.3$\\	
No TSN during training~\ref{abl:TSN}	& $88.2 \pm 1.4$ & $85.8 \pm 1.7$ & $87.1 \pm 1.8$\\
Training from scratch	~\ref{abl:Kin}	& $41.0 \pm 3.5$ & $37.6 \pm 3.5$ & $34.2 \pm 5.8$\\	
3D ConvNet ($K = 5$)		& $93.1 \pm 2.8$ & $92.3 \pm 3.0$ & $92.4 \pm 2.9$ \\
\noalign{\smallskip}\hline
\end{tabular}
\end{table}

To evaluate the impact of our main design decisions on skill classification performance, 
we investigate the following questions:
\begin{enumerate}[(a)]
	\item What is the benefit of using a 3D~ConvNet? 	
	\item What is the benefit of extending the 3D~ConvNet into a Temporal Segment Network during training?		
	\item What is the benefit of pretraining the 3D~ConvNet on the Kinetics dataset?
\end{enumerate}
To this end, we re-run our experiments with one of the following modifications. Unless stated otherwise, we stick to the training procedure described in section~\ref{sec:methods}.
We restrict the experiments to the knot tying task, which had turned out to be most challenging. As input modality, we use optical flow.
\begin{enumerate}[(a)]
	\item We use the 2D~ConvNet \emph{Inception-v3}~\cite{Szegedy2016} instead of Inception-v1~I3D. This also requires changing the video snippet structure: we use stacks of 5 dense optical flow fields calculated at a frequency of 5~Hz. The 2D~ConvNet is initialized with the weights of an Inception-v3 TSN model\footnote{\url{http://yjxiong.me/others/kinetics_action}}, pretrained on Kinetics. For finetuning, the weights of all layers up to, but not including, the last Inception block (\texttt{Mixed\_10}) are frozen.
		\label{abl:3D}
	\item During training, we do not extend the 3D~ConvNet into a TSN. Instead, we train the network on single video snippets that are sampled at random temporal positions from the video. This equals setting the number of segments $K$ to~1. Because the network sees only one snippet, instead of~10, from each video during one epoch of training, we multiply the number of training epochs by factor~$10$. 
		\label{abl:TSN}
	\item Instead of initializing the 3D~ConvNet with weights obtained by training on the Kinetics dataset, we train the TSN from scratch. For initialization, we use PyTorch's standard routine and initialize each layer with weights sampled uniformly from the range $(-\sqrt{\frac{1}{n}}, \sqrt{\frac{1}{n}})$. Here, $n$ denotes the total number of weights in the layer. 
Because we need to train all layers of the network, more computational resources are required. In fact, we cannot fit the model with $K = 10$ segments on the GPU.
Thus, we train the TSN with $K = 5$ segments and a batch size of~1.
Because we halve the number of video snippets that are processed per epoch, we double the number of training epochs.
	
	For fair comparison, we conduct one further experiment where we use a pretrained 3D~ConvNet, as proposed, but train the TSN also with only $K = 5$ segments, a batch size of~1, and for 2400~training epochs.
		\label{abl:Kin}
\end{enumerate}
The LOSO cross-validation results of the ablation experiments can be found in table~\ref{tab:abl}. The evaluation metrics are calculated as described in section~\ref{ss:loso} and are averaged over four evaluation runs.

\subsection{LOUO evaluation}

\begin{figure*}
  \includegraphics[width=0.36\textwidth]{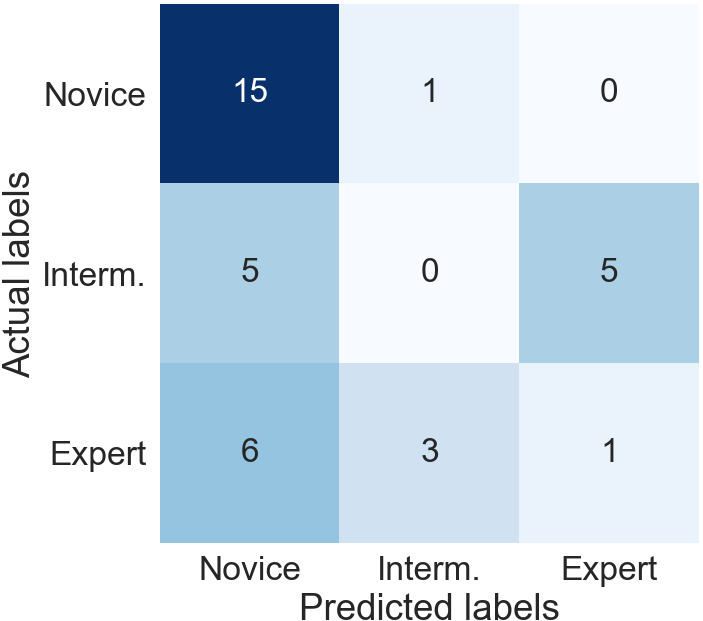}
  \centering
 \caption{Predictions on the knot tying task using the optical flow modality, obtained from one LOUO evaluation run}
 \label{fig:LOUO}       
\end{figure*}

Establishing LOUO validity is important to demonstrate that a skill assessment algorithm will perform well on new subjects.
However, we believe that the data provided in JIGSAWS is too limited to produce meaningful LOUO cross-validation results for learning-based skill assessment approaches. 

One problem is that there are only two experts and two intermediates amongst the participants in JIGSAWS. 
Thus, the  training data is not representative for many of the LOUO splits.
For example, when the test split consists of performances by an expert, the corresponding
training data contains examples of only one other expert,
exhibiting a severe lack of intra-class variability.
This issue can  be observed in figure~\ref{fig:LOUO}, which illustrates the predictions on the knot tying task obtained by a model that was trained following the LOUO cross-validation scheme. While the model classifies most performances of novices correctly, it struggles to recognize performances of intermediates and experts as expected.

\section{Discussion}
\label{sec:discussion}

As can be seen in tables~\ref{tab:1} and~\ref{tab:2}, our method achieves high classification accuracies that are on a par with state-of-the-art approaches. 
Both evaluated input modalities, RGB and optical flow, perform similarly well. There may be a slight advantage in using optical flow because this achieves 100\% classification accuracy on the needle passing task.
Further exploring the strengths and weaknesses of both modalities as well as possibilities to combine them is left for future work.
All in all, our model can learn to extract discriminative patterns from video data that distinguish experienced surgeons from less experienced ones. 

Notably, our method succeeds to learn from highly complex video data while the other approaches analyze the robot kinematics, which consist of only 76~variables.
The ability to analyze raw video 
opens up new application areas. For example, our method could be used on low-cost surgical trainers, which lack sophisticated data acquisition systems.

In line with the other approaches, our method achieves the lowest classification accuracy on the knot tying task. Here, our model mostly misclassifies intermediate surgeons as novices (see figure~\ref{fig:CM1}). 
However, we want to point out that the skill labels in JIGSAWS were assigned to participants based on their hours of robotic surgical experience (novice: $< 10$ hours, expert: $> 100$ hours, intermediate: in between). Hence, it might be possible that intermediate surgeons still exhibit novice-like motion patterns when performing the challenging task of knot tying, leading to a justified misclassification by our model.

It is noteworthy that both our 3D~ConvNet and the ConvNet described by Ismail Fawaz et al.~\cite{Fawaz2018} outperform the ConvNet proposed by Wang and Fey~\cite{Wang2018}. 
One reason might be that \cite{Wang2018} use a rather generic network architecture while \cite{Fawaz2018} draw from domain knowledge to design their network specifically for the skill classification task.
Additionally, the ConvNet by \cite{Fawaz2018} processes the complete time series of kinematic variables at once, while the ConvNet by \cite{Wang2018} only regards short snippets with a  duration of up to 3~seconds. Furthermore, \cite{Wang2018} make the simplifying assumption that the skill label of a snippet is the same as the label assigned to the  video containing that snippet.
In contrast, we extend our ConvNet into a Temporal Segment Network to avoid explicit assignment of skill labels to invidual snippets.

The handcrafted feature based on approximate entropy (ApEn)~\cite{Zia2018} still outperforms our approach. 
However, we believe that learning-based methods have more potential to solve advanced problems for computer-assisted surgical training. 
One example would be the automatic detection of procedural errors, which cannot be detected based on dexterity features alone.

The results of the ablation studies (see table~\ref{tab:abl}) indicate that our method benefits from the combination of a 3D~ConvNet architecture with TSN-based training.
Using a 3D~ConvNet enables spatiotemporal feature learning from video. Training the 3D~ConvNet as TSN allows the network to look at multiple snippets of a video at a time, finding consensus to resolve contradictions and ambiguities of single snippets. 

Pretraining the 3D~ConvNet on the Kinetics dataset turned out to be the most important step. Indeed, a TSN that is trained from scratch achieves only 41\%  classification accuracy on the knot tying task. 
This drastic drop in performance cannot be explained by the circumstance that, due to memory limits, the TSN is trained using only $K = 5$ segments. 
In fact, we observed a classical overfitting effect where the model had converged and fit the training data perfectly, but could not generalize to unseen examples. 

The proposed approach still has its limitations. First, our model can only detect motion patterns that can be observed in rather short video snippets. It cannot explicitly model long-term temporal relations in surgical video. 
Secondly, as a deep learning-based method, our model is rather data hungry and will not work without a comprehensive video data base of surgical task performances that exhibit varying levels of surgical skill. Fortunately, it is relatively easy to record video of traditional or robot-assisted surgery or surgical training. Thus, we are optimistic that enough data can be collected and annotated to establish LOUO validity in future work. 
Another approach to address data limitations could be the investigation of semi- and self-supervised learning algorithms.

\section{Conclusion}
\label{sec:conclusion}
In this paper, we describe a deep learning-based approach to assess technical surgical skill using video data only. 
Using the Temporal Segment Network (TSN) framework, we finetune a pretrained 3D~ConvNet on stacks of video frames or optical flow fields for the task of surgical skill classification. 

Our method achieves competitive video-level classification accuracies of at least 95\% on the JIGSAWS dataset, which contains recordings of basic tasks in robot-assisted minimally invasive surgery.
This indicates that the 3D~ConvNet is able to learn meaningful patterns from the data, which enable the differentiation between expert, intermediate, and novice surgeons.  

Future work will investigate how to integrate RGB video data and optical flow into one framework to fully exploit the richness of video data for surgical skill assessment.
The ultimate goal is to develop a thorough understanding of executions of surgical tasks, where we not only model surgical motion or dexterity but also the interactions between surgical actions and the environment
as well as long-range temporal structure.
A crucial step will be the extension of the datasets currently available for surgical skill evaluation
in order to facilitate large-scale training and evaluation of deep neural networks. Especially the curation and annotation of collected video data will require a close collaboration with surgical educators.

\begin{acknowledgements}
The authors would like to thank the Helmholtz-Zentrum Dresden-Rossendorf (HZDR) for granting access to their GPU cluster for running additional experiments during paper revision.
\end{acknowledgements}

\small {
\noindent
\textbf{Conflict of Interest:} The authors, Isabel Funke, S\"oren Torge Mees, J\"urgen Weitz, and Stefanie Speidel, declare that they have no conflict of interest.

\medskip \noindent
\textbf{Ethical approval:} This article does not contain any studies with human or animal subjects performed by any of the authors.

\medskip \noindent
\textbf{Informed consent:} This articles does not contain patient data.
}

\bibliographystyle{spmpsci}      

\bibliography{references}   

\begin{thebibliography}{10}
\providecommand{\url}[1]{{#1}}
\providecommand{\urlprefix}{URL }
\expandafter\ifx\csname urlstyle\endcsname\relax
  \providecommand{\doi}[1]{DOI~\discretionary{}{}{}#1}\else
  \providecommand{\doi}{DOI~\discretionary{}{}{}\begingroup
  \urlstyle{rm}\Url}\fi

\bibitem{Ahmed2011}
Ahmed, K., Miskovic, D., Darzi, A., Athanasiou, T., Hanna, G.B.: Observational
  tools for assessment of procedural skills: a systematic review.
\newblock Am J Surg \textbf{202}(4), 469--480 (2011)

\bibitem{Ahmidi2017}
Ahmidi, N., Tao, L., Sefati, S., Gao, Y., Lea, C., Haro, B.B., Zappella, L.,
  Khudanpur, S., Vidal, R., Hager, G.D.: A dataset and benchmarks for
  segmentation and recognition of gestures in robotic surgery.
\newblock IEEE Trans Biomed Eng \textbf{64}(9), 2025--2041 (2017)

\bibitem{Bouget2017}
Bouget, D., Allan, M., Stoyanov, D., Jannin, P.: Vision-based and marker-less
  surgical tool detection and tracking: a review of the literature.
\newblock Med Image Anal \textbf{35}, 633--654 (2017)

\bibitem{opencv_library}
Bradski, G.: {The OpenCV Library}.
\newblock Dr. Dobb's Journal of Software Tools  (2000)

\bibitem{Bromley1994}
Bromley, J., Guyon, I., LeCun, Y., S{\"a}ckinger, E., Shah, R.: Signature
  verification using a "siamese" time delay neural network.
\newblock In: NIPS, pp. 737--744 (1994)

\bibitem{Carreira2017}
Carreira, J., Zisserman, A.: Quo vadis, action recognition? {A} new model and
  the {Kinetics} dataset.
\newblock In: CVPR, pp. 4724--4733 (2017)

\bibitem{Chmarra2007}
Chmarra, M.K., Grimbergen, C.A., Dankelman, J.: Systems for tracking minimally
  invasive surgical instruments.
\newblock Minim Invasive Ther Allied Technol \textbf{16}(6), 328--340 (2007)

\bibitem{Doughty2017}
Doughty, H., Damen, D., Mayol{-}Cuevas, W.W.: Who's better, who's best: Skill
  determination in video using deep ranking.
\newblock In: CVPR, pp. 6057 -- 6066 (2018)

\bibitem{Du2018}
Du, X., Kurmann, T., Chang, P.L., Allan, M., Ourselin, S., Sznitman, R., Kelly,
  J.D., Stoyanov, D.: Articulated multi-instrument {2D} pose estimation using
  fully convolutional networks.
\newblock IEEE Trans Med Imaging \textbf{37}(5), 1276--1287 (2018)

\bibitem{Fard2018}
Fard, M.J., Ameri, S., Darin~Ellis, R., Chinnam, R.B., Pandya, A.K., Klein,
  M.D.: Automated robot-assisted surgical skill evaluation: Predictive
  analytics approach.
\newblock Int J Med Robot \textbf{14}(1), e1850 (2018)

\bibitem{Gao2014}
Gao, Y., Vedula, S.S., Reiley, C.E., Ahmidi, N., Varadarajan, B., Lin, H.C.,
  Tao, L., Zappella, L., B{\'e}jar, B., Yuh, D.D., Chen, C.C.G., Vidal, R.,
  Khudanpur, S., Hager, G.D.: {JHU-ISI} gesture and skill assessment working
  set ({JIGSAWS}): A surgical activity dataset for human motion modeling.
\newblock In: M2CAI (2014)

\bibitem{goh2015}
Goh, A.C., Aghazadeh, M.A., Mercado, M.A., Hung, A.J., Pan, M.M., Desai, M.M.,
  Gill, I.S., Dunkin, B.J.: Multi-institutional validation of fundamental
  inanimate robotic skills tasks.
\newblock The Journal of Urology \textbf{194}(6), 1751--1756 (2015)

\bibitem{He2016}
He, K., Zhang, X., Ren, S., Sun, J.: Deep residual learning for image
  recognition.
\newblock In: CVPR, pp. 770--778 (2016)

\bibitem{Fawaz2018}
Ismail~Fawaz, H., Forestier, G., Weber, J., Idoumghar, L., Muller, P.A.:
  Evaluating surgical skills from kinematic data using convolutional neural
  networks.
\newblock In: MICCAI, pp. 214--221 (2018)

\bibitem{Ji2013}
Ji, S., Xu, W., Yang, M., Yu, K.: {3D} convolutional neural networks for human
  action recognition.
\newblock IEEE Trans Pattern Anal Mach Intell \textbf{35}(1), 221--231 (2013)

\bibitem{Jin2018}
Jin, A., Yeung, S., Jopling, J., Krause, J., Azagury, D., Milstein, A.,
  Fei-Fei, L.: Tool detection and operative skill assessment in surgical videos
  using region-based convolutional neural networks.
\newblock In: WACV, pp. 691--699 (2018)

\bibitem{Kay2017}
Kay, W., Carreira, J., Simonyan, K., Zhang, B., Hillier, C., Vijayanarasimhan,
  S., Viola, F., Green, T., Back, T., Natsev, P., Suleyman, M., Zisserman, A.:
  The kinetics human action video dataset.
\newblock arXiv preprint arXiv:1705.06950  (2017)

\bibitem{Kingma2015}
Kingma, D.P., Ba, J.: Adam: {A} method for stochastic optimization.
\newblock In: ICLR (2015)

\bibitem{Laina2017}
Laina, I., Rieke, N., Rupprecht, C., Vizca{\'\i}no, J.P., Eslami, A., Tombari,
  F., Navab, N.: Concurrent segmentation and localization for tracking of
  surgical instruments.
\newblock In: MICCAI, pp. 664--672 (2017)

\bibitem{Lecun2015}
LeCun, Y., Bengio, Y., Hinton, G.: Deep learning.
\newblock Nature \textbf{521}(7553), 436--444 (2015)

\bibitem{Martin1997}
Martin, J., Regehr, G., Reznick, R., Macrae, H., Murnaghan, J., Hutchison, C.,
  Brown, M.: Objective structured assessment of technical skill ({OSATS}) for
  surgical residents.
\newblock British journal of surgery \textbf{84}(2), 273--278 (1997)

\bibitem{Paszke2017}
Paszke, A., Gross, S., Chintala, S., Chanan, G., Yang, E., DeVito, Z., Lin, Z.,
  Desmaison, A., Antiga, L., Lerer, A.: Automatic differentiation in {PyTorch}.
\newblock In: NIPS Workshops (2017)

\bibitem{peters2004}
Peters, J.H., Fried, G.M., Swanstrom, L.L., Soper, N.J., Sillin, L.F.,
  Schirmer, B., Hoffman, K., {Sages FLS Committee}: Development and validation
  of a comprehensive program of education and assessment of the basic
  fundamentals of laparoscopic surgery.
\newblock Surgery \textbf{135}(1), 21--27 (2004)

\bibitem{Simonyan2014}
Simonyan, K., Zisserman, A.: Two-stream convolutional networks for action
  recognition in videos.
\newblock In: NIPS, pp. 568--576 (2014)

\bibitem{Szegedy2016}
Szegedy, C., Vanhoucke, V., Ioffe, S., Shlens, J., Wojna, Z.: Rethinking the
  inception architecture for computer vision.
\newblock In: CVPR, pp. 2818--2826 (2016)

\bibitem{Tao2012}
Tao, L., Elhamifar, E., Khudanpur, S., Hager, G.D., Vidal, R.: Sparse hidden
  markov models for surgical gesture classification and skill evaluation.
\newblock In: IPCAI, pp. 167--177 (2012)

\bibitem{Tran2015}
Tran, D., Bourdev, L., Fergus, R., Torresani, L., Paluri, M.: Learning
  spatiotemporal features with 3d convolutional networks.
\newblock In: ICCV, pp. 4489--4497 (2015)

\bibitem{Vedula2017}
Vedula, S.S., Ishii, M., Hager, G.D.: Objective assessment of surgical
  technical skill and competency in the operating room.
\newblock Annu Rev Biomed Eng \textbf{19}, 301--325 (2017)

\bibitem{Wang2018TSN}
Wang, L., Xiong, Y., Wang, Z., Qiao, Y., Lin, D., Tang, X., Gool, L.V.:
  Temporal segment networks for action recognition in videos.
\newblock IEEE Trans Pattern Anal Mach Intell  (2018)

\bibitem{Wang2016}
Wang, L., Xiong, Y., Wang, Z., Qiao, Y., Lin, D., Tang, X., Van~Gool, L.:
  Temporal segment networks: Towards good practices for deep action
  recognition.
\newblock In: ECCV, pp. 20--36. Springer (2016)

\bibitem{Wang2018}
Wang, Z., Majewicz~Fey, A.: Deep learning with convolutional neural network for
  objective skill evaluation in robot-assisted surgery.
\newblock Int J Comput Assist Radiol Surg  (2018)

\bibitem{Zach2007}
Zach, C., Pock, T., Bischof, H.: A duality based approach for realtime {TV-L1}
  optical flow.
\newblock In: Joint Pattern Recognition Symposium, pp. 214--223. Springer
  (2007)

\bibitem{Zia2018}
Zia, A., Essa, I.: Automated surgical skill assessment in {RMIS} training.
\newblock Int J Comput Assist Radiol Surg \textbf{13}(5), 731--739 (2018)

\bibitem{Zia2018Video}
Zia, A., Sharma, Y., Bettadapura, V., Sarin, E.L., Essa, I.: Video and
  accelerometer-based motion analysis for automated surgical skills assessment.
\newblock Int J Comput Assist Radiol Surg \textbf{13}(3), 443--455 (2018)

\end{thebibliography}

\end{document}